\documentclass[10pt,twocolumn,letterpaper]{article}

\usepackage{wacv}
\usepackage{times}
\usepackage{epsfig}
\usepackage{graphicx}
\usepackage{amsmath}
\usepackage{amssymb}
\usepackage{xcolor}
\usepackage{verbatim}
\usepackage{algorithm}
\usepackage{multirow}
\usepackage{dsfont}
\usepackage{caption}
\usepackage{footnote}
\makesavenoteenv{tabular}
\makesavenoteenv{table}
\usepackage[noend]{algpseudocode}
\graphicspath{ {./} }
\definecolor{darkspringgreen}{rgb}{0.09, 0.45, 0.27}
\wacvfinalcopy % *** Uncomment this line for the final submission

\def\wacvPaperID{299} % *** Enter the wacv Paper ID here

\ifwacvfinal\fi
\begin{document}

%%%%%%%%% TITLE
\title{Hybrid Binary Networks: Optimizing for Accuracy, Efficiency and Memory}

% Authors at the same institution
%\author{First Author \hspace{2cm} Second Author \\
% Authors at different institutions
%\author{Ameya Prabhu, Vishal Batchu, Rohit Gajawada, Sri Aurobindo Munagala, and Anoop Namboodiri\\
\author{Ameya Prabhu\quad
Vishal Batchu\quad
Rohit Gajawada\quad
Sri Aurobindo Munagala\quad
Anoop Namboodiri\\
Center for Visual Information Technology, Kohli Center on Intelligent Systems\\
IIIT-Hyderabad, India\\
{\tt\small \{ameya.prabhu@research., vishal.batchu@students., rohit.gajawada@students.},\\{\tt\small s.munagala@research., anoop@\}iiit.ac.in}}

\maketitle
\ifwacvfinal\thispagestyle{empty}\fi

%%%%%%%%% ABSTRACT
\begin{abstract}
  \noindent Binarization is an extreme network compression approach that provides large computational speedups along with energy and memory savings, albeit at significant accuracy costs. We investigate the question of where to binarize inputs at layer-level granularity and show that selectively binarizing the inputs to specific layers in the network could lead to significant improvements in accuracy while preserving most of the advantages of binarization. We analyze the binarization tradeoff using a metric that jointly models the input binarization-error and computational cost and introduce an efficient algorithm to select layers whose inputs are to be binarized. Practical guidelines based on insights obtained from applying the algorithm to a variety of models are discussed.\\ Experiments on Imagenet dataset using AlexNet and ResNet-18 models show 3-4\% improvements in accuracy over fully binarized networks with minimal impact on compression and computational speed. The improvements are even more substantial on sketch datasets like TU-Berlin, where we match state-of-the-art accuracy as well, getting over 8\% increase in accuracies. We further show that our approach can be applied in tandem with other forms of compression that deal with individual layers or overall model compression (e.g., SqueezeNets). Unlike previous quantization approaches, we are able to binarize the weights in the last layers of a network, which often have a large number of parameters, resulting in significant improvement in accuracy over fully binarized models.
\end{abstract}

\vspace{-0.25cm}
\section{Introduction}

\noindent Convolutional Neural Networks (CNNs) have found applications in many vision-related domains ranging from generic image-understanding for self-driving cars \cite{bojarski2016end} and automatic image captioning \cite{you2016image,johnson2016densecap} to recognition of specific image parts for scene-text recognition \cite{mishra2012top,neumann2012real} and face-based identification \cite{taigman2014deepface}.
\begin{figure}[t]
%\resizebox{0.5\textwidth}{!}{
\centering
\includegraphics[scale=0.4]{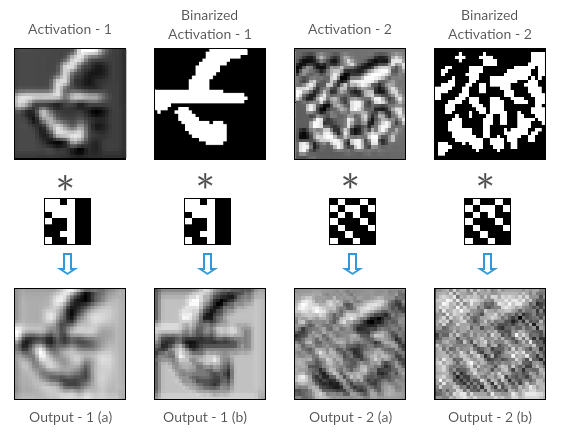}
%}
\caption{Convolution of binary and non-binary activations of two different layers. Note that the error introduced due to binarization is minimal in the first pair compared to the second. Hence, efficiently deciding \textit{which} layers to binarize could contribute significantly to the overall accuracy of the network and not damage the speed-ups.}
\label{fig:introdiag}
\vspace{-0.5cm}
\end{figure}

After the introduction of AlexNet \cite{alex2012alexnet}, several architectural improvements were proposed to push image recognition accuracy, such as VGG-Net \cite{simonyan2014very}, but these models were massive both in terms of memory usage and computational costs. AlexNet has around 60 million parameters in the network, while VGG has around 138 million, requiring 1.5 billion FLOPs and 19.6 billion FLOPs respectively for inference. The computational requirements make these architectures inappropriate for smaller portable systems such as mobiles and other embedded systems. These networks also use large amounts of energy, creating a bottleneck for performance improvements. Full-precision multiply-accumulate (MAC) operations in convolutional layers consume 30x more power than integer MAC operations (see Table \ref{table:mac-energy}). 

Since these applications would be deployed on resource-constrained systems, CNN compression is an important emerging area for research on vision applications \cite{hubara2016quantized,zhou2016dorefa, han2015deep, liu2017learning, moczulski2015acdc, yang2015deep, he2016deep, iandola2016squeezenet}. One of the methods of compression: Quantization, can help networks consume far less power, memory, and incur lower computational costs.

\begin{table}[t]
\begin{center}
\begin{tabular}{|l|c|c|c|c|}
\hline
{\bf Operation} & {\bf MUL} & {\bf Power} & {\bf ADD} & {\bf Power}\\ 
\hline
32-bit Float & 3.7pJ & 18.5x & 0.9pJ & 30x \\
16-bit Float & 1.1pJ & 5.5x & 0.4pJ &  13.3x\\
8-bit Integer & 0.2pJ & 1x & 0.03pJ & 1x \\
\hline
\end{tabular}
\end{center}
\caption{As shown by Horowitz \etal\cite{horowitz2014power}, power consumption for various operations at 45nm 0.9V. Observe that 8-bit integers require significantly less energy than their equivalent 32-bit floating point operations.}
\label{table:mac-energy}
\end{table}
Quantization has proven to be a powerful compression strategy. Our paper is based on the most extreme form of quantization - Binarization. There are many benefits to binarizing a network. Primarily, having binary weights/activations enables us to use xnor and popcount operations to calculate weighted sums of the inputs to a layer as compared to full-precision multiply-accumulate operations (MACs). This results in significant computational speedup compared to other compression techniques. Secondly, as each binary weight requires only a single bit to represent, one can achieve drastic reductions in run-time memory requirements. Previous research \cite{rastegari2016xnor,hubara2016quantized} shows that it is possible to perform weight and input binarization on large networks with up to 58x speedups and 10.4x compression ratios, albeit with significant drops in accuracy.

In this paper, we explore the problem of hybrid binarization of a network. We propose a technique devised from our investigation into the question as to {\it where and which quantities of a network should one binarize}, with respect to inputs to a layer - to the best of our knowledge, this is the first work that explores this question. We observe in Figure \ref {fig:introdiag} that in a trained fully binarized model, binarization in certain layers induces minimal error, whereas in others, the error obtained is significant. Our proposed partition algorithm, when run on trained fully binarized models can design effective architectures. When these hybrid models are trained from scratch, they  achieve a balance between compression, speedup, energy-efficiency, and accuracy, compared to fully binarized models. We conduct extensive experiments applying our method to different model architectures on popular large-scale classification datasets over different domains. The resulting models achieve significant speedups and compression with significant accuracy improvements over a fully binarized network.

Our main contribution includes:\vspace{-0.2cm}
\begin{enumerate}
\item A metric to jointly optimize binarization-errors of layers and the associated computational costs;\vspace{-0.2cm}
\item A partitioning algorithm to find suitable layers for input binarization, based on the above metric, which generates hybrid model architectures which if trained from scratch, achieve a good balance between compression, speedup, energy-efficiency, and accuracy;\vspace{-0.2cm}
\item Insights into what the algorithm predicts, which can provide an intuitive framework for understanding why binarizing certain areas of networks give good benefits;\vspace{-0.2cm}
\item Hybrid model architectures for AlexNet, ResNet-18, Sketch-A-Net and SqueezeNet with over 5-8\% accuracy improvements on various datasets; and\vspace{-0.2cm}
\item A demonstration that our technique that achieves significant compression in tandem with other compression methods.
\end{enumerate}\vspace{-0.2cm}

{\bf Reproducibility:} Our implementation can be found on GitHub \footnote{https://github.com/erilyth/HybridBinaryNetworks-WACV18}.

\section{Related Work}

CNNs are often over-parametrized with high amounts of redundancy, increasing memory costs and making computation unnecessarily expensive. Several methods were proposed to compress networks and eliminate redundancy, which we summarize below.

{\bf Space-efficient architectures}: Designing compact architectures for deep networks helps save memory and computational costs. Architectures such as ResNet \cite{he2016deep}, DenseNet \cite{huang2017densely} significantly reduced model size compared to VGG-Net by proposing a bottleneck structure to reduce the number of parameters while improving speed and accuracy. SqueezeNet \cite{iandola2016squeezenet} was another model architecture that achieved AlexNet-level accuracy on ImageNet with 50x fewer parameters by replacing 3x3 filters with 1x1 filters and late downsampling in the network. MobileNets \cite{howard2017mobilenets} and ShuffleNets \cite{zhang2017shufflenet} used depthwise separable convolutions to create small models, with low accuracy drop on ImageNet.

{\bf Pruning and Quantization}: Optimal Brain Damage \cite{Cunn1} and Optimal Brain Surgeon \cite{Hassibi} used the Hessian of the loss function to prune a network by reducing the number of connections. Deep Compression \cite{han2015deep} reduced the number of parameters by an order of magnitude in several state-of-the-art neural networks through pruning. It further reduced non-runtime memory by employing trained quantization and Huffman coding. Network Slimming \cite{liu2017learning} took advantage of channel-level sparsity in networks, by identifying and pruning out non-contributing channels during training. HashedNets \cite{chen2015hashed} performed binning of network weights using hash functions. INQ \cite{zhou2017inq} used low-precision 16 bit-quantized weights and achieved an 8x reduction in memory consumption, using 4 bits to represent 16 distinct quantized values and 1 bit to represent zeros specifically.

{\bf Binarization:}  BinaryConnect \cite{courbariaux2015binaryconnect} obtained huge compression in CNNs where all weights had only two allowed states (+1, -1) using Expectation Back Propagation (EBP). Approaches like \cite{hubara2016quantized, li2016ternary, zhu2016trained} train deep neural networks using low precision multiplications, bringing down memory required drastically, showing that these models could be fit on memory constrained devices. DoReFa-net \cite{zhou2016dorefa} applied low bit width gradients during back-propagation. XNOR-Net \cite{rastegari2016xnor} multiplied binary weights and activations with scaling constants based on layer norms. QNNs \cite{hubara2016quantized} extended BNNs\cite{courbariaux2016binarized}, the first method using binary weights and inputs to successfully achieve accuracy comparable to their corresponding 32-bit versions on constrained datasets using higher bit quantizations. HWGQ-Net \cite{cai2017deep} introduces a better suited activation function for binary networks. HTCBN \cite{tang2017train} introduce helpful techniques such as replacing ReLU layers with PReLU layers and a scale layer to recover accuracy loss on binarizing the last layer, to effectively train a binary neural network. Hou \etal \cite{hou2016loss} use Hessian approximations to minimize loss w.r.t binary weights during training. Anderson \etal \cite{anderson2017high} offers a theoretical analysis of the workings of binary networks, in terms of high-dimensional geometry.

Unlike previous works in this area, we look at binarizing specific parts of a network, instead of simply binarizing the inputs to all the layers end-to-end. We see in later sections, binarizing the right areas in the network contributes significantly to the overall accuracy of the network and does not damage its speed-ups.

\section{Hybrid Binarization}

We define certain conventions to be used throughout the paper. We define a WBin CNN to be a CNN having the weights of convolutional layers binarized (referred to as WeightBinConv layers), FBin CNN to be a CNN having both inputs and weights of convolutional layers binarized (referred to as FullBinConv layers) and FPrec CNN to be the original full-precision network having both weights and inputs of convolutional layers in full-precision (referred to as Conv layers). We compare the FBin and WBin networks with FPrec networks at specific layers.
%Since XNOR-Net achieves a 58x times speedup, we assume roughly 58 binary operations are happening effectively like a FLOP operation.
%The details on FLOP calculations in binary networks are presented in detail in section x.

Table \ref{table:versions_imagenet} and Table \ref{table:tub_recacc} in the Experiments section show test accuracies for WBin, FBin and FPrec networks of different models. Observe that there is very little loss in accuracy from FPrec to WBin networks with significant memory compression and fewer FLOPs. However, as we go from WBin to FBin networks, there is a significant drop in accuracy along with the trade-off of significantly lower FLOPs in FBin over WBin networks. %The primary focus has been to improve the accuracies of FBin networks - with approaches either developing better binarization techniques (such as XNOR-Net) or discarding input binarization altogether in favor of multi-bit quantization, as seen in the more recent works after XNOR-Net. The problem with those approaches is discarding input binarization which has explicitly optimized algorithms for convolution using XNOR and Popcount operations and using normal operations, leads to much slower convolution algorithms - leading to massive increases in the number of FLOPs.
Hence, we focus on improving the accuracies of FBin networks along with preserving the lower FLOPs as far as possible by investigating which activations to binarize.

\begin{figure*}[t]
\vspace*{-0.50cm}
\resizebox{\textwidth}{!}{
\includegraphics[]{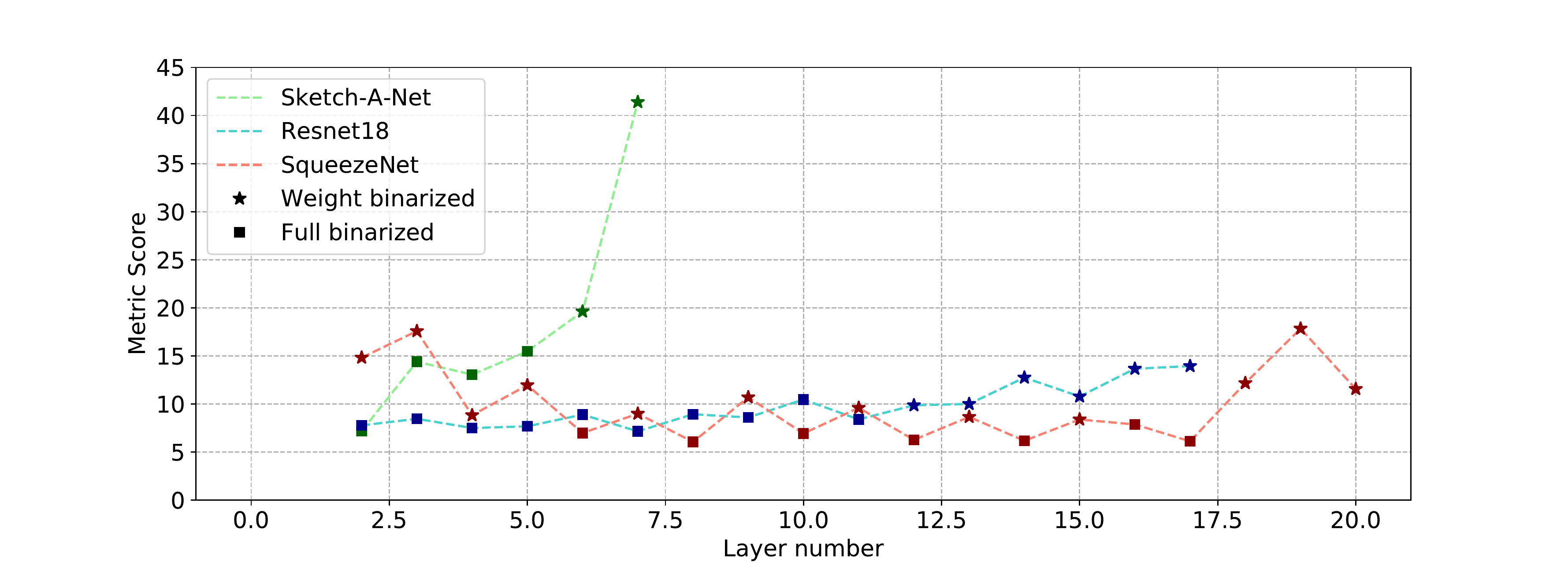}
}
\vspace*{-0.5cm}
\caption{Binarization-error metric across layers for Sketch-A-Net, ResNet-18, and SqueezeNet. Stars indicate that the layer was replaced with a WeightBinConv layer, while squares indicate the FullBinConv layer was retained in the FBin model. We see that the algorithm selects the last layers in the case of Sketch-A-Net and ResNet, while in the case of SqueezeNet, it selects the first four, last three and some alternate intermediate layers to be replaced by WeightBinConv layers, retaining the rest  as FullBinConv layers.}
\label{fig:metric}
\vspace*{-0.4cm}
\end{figure*}

\subsection{Error Metric: Optimizing Speed \& Accuracy}
Full-precision inputs $\mathbf{I} \in \mathbb{R}^n$, are approximated by binary matrix $\mathbf{I_B} \in \mathbb\{-1,+1\}^n$. The optimal binary representation $\mathbf{I_B}$ is calculated by
\begin{equation}\mathbf{I_B}^\ast  = argmin(\parallel\mathbf{I}-\mathbf{I_B}\parallel^2)\end{equation}
XNOR-Net\cite{rastegari2016xnor} minimized the error function:
\begin{equation} \mathbf{E} = \frac{\parallel \mathbf{I} - \mathbf{I_B} \parallel^2}{n}\end{equation}
In order to do that, they maximized $\mathbf{I}^\top\mathbf{I_B}$ and proposed the binary activation $\mathbf{I_B}$ to be 
\begin{equation}\mathbf{I_B}^\ast = \underset{\mathbf{\mathbf{I_B}}}{\mathrm{argmax}}(\mathbf{I}^\top\mathbf{I_B}), \mathbf{I_B} \in \{-1,+1\}^n , \mathbf{I} \in \mathbb{R}^n \end{equation}, obtaining the optimal $\mathbf{I_B}^\ast$ can be shown to be  $sgn(\mathbf{I}).$
%Similarly (writesomemore here), $ \thinspace \mathbf{W_B}^\ast = sgn(\mathbf{W})$
%\begin{equation} \mathbf{I} \ast \mathbf{W} \approx \mathbf{I_B} \ast \mathbf{W_B} \approx Bitcount(sgn(\mathbf{I}) \oplus sgn(\mathbf{W}) ) \label{binconvequation} \end{equation} 

We need to investigate {\it where} to replace FullBinConv with WeightBinConv layers. In order to optimize for accuracy, we need to measure the efficacy of the binary approximation for inputs to any given layer. A good metric of this is the average error function calculated over a subset of training images $\mathbf{E}$ (defined in Eq. 2) used to calculate the optimal $I_B$ itself, which is explicitly being minimized in the process. Hence, we use that error function to capture the binarization error.

Similarly to optimize speed, we need to convert layers with low number of FLOPs to WeightBinConv and layers having high number of FLOPs should be kept in FullBinConv. Since we need to jointly optimize both, we propose a metric that tries to achieve a good tradeoff between the two quantities. A simple but effective metric is the linear combination \begin{equation} \mathbf{M} = \mathbf{E} + \gamma \cdot \frac{1}{\mathbf{NF}} \end{equation} where $\gamma$ is the tradeoff ratio, $\mathbf{NF}$ is the number of flops in the layer and $\mathbf{E}$ is the binarization error per neuron. The trade-off ratio $\gamma$ is a hyperparameter which ensures that both the terms are of comparable magnitude. Figure \ref{fig:metric}, captures the layer-wise variation of the error metric across multiple models.

\subsection{Partitioning Algorithm}

We aim to partition the layers of a network into two parts, one set of layers to keep FullBinConv and the other set which are replaced with WeightBinConv layers. A naive but intuitive partitioning algorithm would be to sort the list of metric errors {\bf $M$} and replace FullBinConv layers which have highest error values {\bf $M_i$} one-by-one with WeightBinConv layers, train new hybrid models and stop when the accuracies in the retrained models stop improving i.e when the maxima in accuracy v/s flops tradeoff is reached. However, we need a partitioning algorithm which gives informed guesses on where are the effective places to partition the set. This would avoid the long retraining times and large resources required to try every possible option for a hybrid model. We propose a layer selection algorithm that gives informed partitions from a trained FBin model, helping us to determine which layers are to be converted to WeightBinConv and which layers are to be converted to FullBinConv without having to train all possible hybrid models from scratch.

\begin{algorithm}[t]
\caption{{\bf Partition Algorithm} \\Marks layers for binarization and creates a hybrid network.}
\begin{algorithmic}[1]
\State Inputs $\Rightarrow$Layer-wise Binarization Errors
\\

\State \texttt{Initialization}
\State $P$ = Total convolutional layers
\State $R$ = Hybridization Ratio
\State ToConvert = List() \\
\State \texttt{Mark binary layers}
\For{$N$ = 2 to P}
	\State Compute KMeans with $N$ means
	\State $K$ = Number of layers in highest-error cluster
	\If{$K/P \leq R$}
	\For{$Q$ in high-error clusters}
        	\State ToConvert.add($Q$) \Comment{Add layer $Q$ }
    \EndFor
    \State Break
	\EndIf
\EndFor 

\\

\State \texttt{Create Hybrid Network}
\State HybridNet = ()

\State HybridNet.Add(Conv)
\\
\For{$N$ = 2 to P}
        \If{$N$ in ToConvert}
            \State HybridNet.Add(WeightBinConv)
        \Else
        	\State HybridNet.Add(FullBinConv)
        \EndIf
\EndFor
\\
\State Output $\Rightarrow$ HybridNet

\end{algorithmic}
\label{alg:partitionalgo}
%\vspace*{-0.75cm}
\end{algorithm}
\begin{figure*}[t]
\resizebox{\textwidth}{!}{
\includegraphics[]{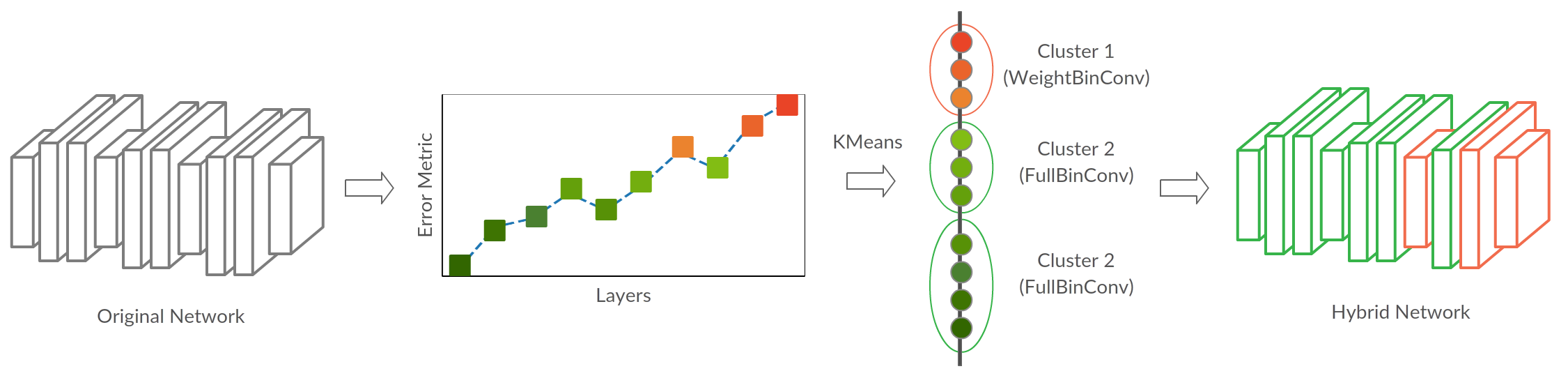}
}
\vspace*{-0.75cm}
\caption{The Procedure: Error metrics from binarization of inputs to the network layers are partitioned into clusters using K-means. The highest error cluster indicates the inputs that are not binarized to generate the hybrid version.}
\label{fig:pipeline}
%\vspace{-1.5cm}
\end{figure*}
Our algorithm starts by taking a trained FBin model. We pass in a subset of the training images and calculate the average error metric for all layers over them. Then we perform K-Means Clustering on the metric values with each point being the metric error of layers as shown in Figure \ref{fig:metric}. We perform the K-Means Clustering for different values of the number of clusters. We find a suitable number of clusters such that the ratio of layers in the highest-error cluster ($K$) to the total number of convolutional layers ($P$) is less than a hyperparameter, which we define as the Hybridization Ratio $R$. Layers with terms falling in the highest mean cluster are converted to WeightBinConv, while the ones in all other clusters are left as FullBinConv. A flow of the algorithm is illustrated in Figure \ref{fig:pipeline} and is explained step-by-step in Algorithm \ref{alg:partitionalgo}. We show metric scores of various layers for different networks in Figure \ref{fig:metric} and indicate which layers are replaced with WeightBinConv/FullBinConv layers. This algorithm guides in forming the architecture of the hybrid model, which is then trained from scratch obtaining the accuracies given in the tables presented in the Experiment section. Note that this algorithm does not change the configuration of the model; it only converts certain layers to their binarized versions.

To give an intuition of what the Hybridization ratio $R$ means, a low $R$ would indicate we need the number of WeightBinConv layers to be low, ensuring a high asymmetry between errors in WeightBinConv and FullBinConv layers, prioritizing saving computational cost. Conversely, a higher $R$ would prioritize accuracy over computational cost. $R$ was set to be 0.4 for AlexNet and ResNet-18, and 0.6 for Squeezenet. Variation with different values of $R$ is further discussed in the experiments section.

\subsection{Impact on Speed and Energy Use}
\noindent{\bf Computational Speedups}: Convolutional operations are computationally expensive. For each convolution operation between an image $\mathbf{I} \in \mathbb{R}^{c_{in} \times h_I \times w_I}$ and weight $\mathbf{W} \in \mathbb{R}^{c_{out} \times h \times w}$, the number of MAC operations required $N$ are $\approx C_{in}C_{out}N_WN_I$ where $N_W = wh$ and $N_I = w_Ih_I$. According to benchmarks done in XNOR-Net, the current speedup obtained in these operations is 58x after including the overhead induced by computing $\alpha$. Accordingly, in later sections, we take one FLOP through a layer as equivalent to 58 binary operations when weights and inputs are binarized. \\ 
{\bf Exploiting filter repetitions}: The number of unique convolutional binary filters is bounded by the size of the filter \cite{hubara2016quantized}. As most of our intermediate convolutional layers have $3\times3$ filters which only have $2^9$ unique filters, we find that the percentage of unique filters decreases as we go deeper into the network. We can exploit this fact to simply prune filters and use that in calculating speedups for binary networks. More details regarding how the speedup was computed is included in the supplementary material.
%{\bf Model Compression}: Every convolutional layer contains $N_p = C_{in}C_{out}N_W$ parameters. Our networks, since with hybrid binarization are able to keep last layer-weights binarized unlike previous approaches, gain 30x compression, as compared to a 10.4x saving obtained by those networks with better accuracies than them. We conjecture that this is due to the output of the last layer being much more expressive enabling it to take advantage of the lower layer outputs.

%{\bf High Power Efficiency}: Power usage is a problematic factor for improving performance. This initiated several efforts to mitigate this issue. Under 45nm CMOS technology, a 32 bit floating point add consumes 0.9pJ, a 32bit SRAM cache access takes 5pJ, while a 32bit DRAM memory access takes 640pJ, which is 3 orders of magnitude of an add operation. Larger models cannot be stored compactly resulting in expensive DRAM accesses. For example, the forward pass of a 1 billion connection neural network, at 20fps would require (20Hz)(1G)(640pJ) = 12.8W just for DRAM access which is infeasible in a typical portable device. Our methods provide two orders of magnitude improvement over these.

%------------------------------------------------------------------------
\section{Experiments and Results}

We report and compare accuracies, speedups and compression between the FPrec model, different kinds of binarization models (WBin and FBin), and their generated hybrid versions of the same. We also present a detailed comparison of our method with several different compression techniques applied on AlexNet \cite{alex2012alexnet}, ResNet-18 \cite{he2016deep}, Sketch-A-Net \cite{eitz2012hdhso} and SqueezeNet \cite{iandola2016squeezenet}. \\
We empirically demonstrate the effectiveness of hybrid binarization on several benchmark image and sketch datasets. We show that our approach is robust and can generalize to different types of CNN architectures across domains.

\subsection{Datasets and Models}
Binary Networks have achieved accuracies comparable to full-precision networks on limited domain/simplified datasets like CIFAR-10, MNIST, SVHN, but show drastic accuracy losses on larger-scale datasets. To compare with state-of-the-art vision, we evaluate our method on ImageNet\cite{imagenet_cvpr09}. To show the robustness of our approach, we test it on sketch datasets, where models fine-tuned with ImageNet are demonstrably not suitable as shown in\cite{yu2015sketch}. Binary networks might be better suited for sketch data due to its binary nature and sparsity of information in the data. 

{\bf ImageNet:} The benchmark dataset for evaluating image recognition tasks, with over a million training images and 50,000 validation images. We report the single-center-crop validation errors of the final models.

{\bf TU-Berlin:} The TU-Berlin \cite{eitz2012hdhso} sketch dataset is the most popular large-scale free-hand sketch dataset containing sketches of 250 categories, with a human sketch-recognition accuracy of 73.1\% on average. 

{\bf Sketchy:} It is a recent large-scale free-hand sketch dataset containing 75,471 hand-drawn sketches from across 125 categories. This dataset was primarily used to cross-validate results obtained on the TU-Berlin dataset and ensure that our approach is robust to the variation in collection of data.

\begin{table}[t]
\begin{center}
\resizebox{\columnwidth}{!}{
\begin{tabular}{|l|c|c|c|c|l|}
\hline
{\bf Technique} & {\bf Acc-Top1} & {\bf Acc-Top5} & {\bf W/I} & {\bf Mem} & {\bf FLOPs} \\
\hline
\multicolumn{6}{|c|}{\sc { \bf AlexNet}} \\
\hline
BNN & 39.5\% & 63.6\% & 1/1 & 32x & 121 (1x)\\
XNOR & 43.3\% & 68.4\% & 1/1 & 10.4x & {\bf 121 (1x)} \\
Hybrid-1 & 48.6\% & 72.1\% & 1/1 & 10.4x & 174 (1.4x)\\
Hybrid-2 & {\bf 48.2\%} & {\bf 71.9\%} & 1/1 & {\bf 31.6x} & 174 (1.4x)\\
\hline
HTCBN & 46.6\% & 71.1\% & 1/2 & 31.6x & 780 (6.4x)\\
DoReFa-Net & 47.7\% & - & 1/2 & 10.4x & 780 (6.4x)\\
\hline
\multicolumn{6}{|c|}{\sc { \bf Res-Net 18}} \\
\hline
BNN & 42.1\% & 67.1\% & 1/1 & 32x & 134 (1x)\\
XNOR & 51.2\% & 73.2\% & 1/1 & 13.4x & {\bf 134 (1x)} \\
Hybrid-1 & 54.9\% & 77.9\% & 1/1 & 13.4x & 359 (2.7x)\\
Hybrid-2 & {\bf 54.8\%} & {\bf 77.7\%} & 1/1 & {\bf 31.2x} & 359 (2.7x)\\
\hline
HTCBN & 53.6\% & - & 1/2 & 31.2x & 1030 (7.7x)\\
\hline
%\multicolumn{6}{|c|}{\sc { \bf NIN-Net}} \\
%\hline
%XNOR-Net & -\% & -\% & 1/1 & - & - \\
%HTCBN* & 51.4\%* & 75.6\%* & 1/2 & - & - \\
%Ours & -\% & -\% & 1/1 & - & - \\
%\hline
\end{tabular}

}
\end{center}
\vspace{-0.4cm}
\caption{
A detailed comparison of accuracy, memory use, FLOPs with popular benchmark compression techniques on ImageNet. Our hybrid models outperform other 1-bit activation models and perform on par with 2-bit models while having a significantly higher speedup. Hybrid-2 models have the last layer binarized.}
\label{table:imagenet_fullcomp}
\end{table}

We use the standard splits with commonly used hyper-parameters to train our models. Each FullBinConv block was structured as in XNOR-Net (Batchnorm-Activ-Conv-ReLU). Each WeightBinConv and Conv block has the standard convolutional block structure (Conv-Batchnorm-ReLU). Weights of all layers except the first were binarized throughout our experiments unless specified otherwise. Note that FLOPs are stated in millions in all diagrams and sections. All networks are trained from scratch independently. The architecture of the hybrid network once designed does not change during training. Additional details about the datasets, model selection and layer-wise description of each of the hybrid models along with experimental details can be found in the supplementary material.

\subsection{Results}

We compare FBin, WBin, Hybrid and FPrec recognition accuracies across models on ImageNet, TU-Berlin and Sketchy datasets. Note that higher accuracies are an improvement, hence stated in green in the table, while higher FLOPs mean more computational expense, hence are stated in red. W/I indicates the number of bits used for weights and inputs to the layer respectively. Note that in the table, the compression obtained is only due to the weight binarization, while the decrease in effective FLOPs are due to activation binarization.

On the ImageNet dataset in Table \ref{table:versions_imagenet}, hybrid versions of AlexNet and ResNet-18 models outperform their FBin counterparts in top-1 accuracy by 4.1\% and 3.6\% respectively, and around 20x compression for both. We also compare with the results of other compression techniques in Table \ref{table:imagenet_fullcomp}.
\begin{table}[t]
\resizebox{\columnwidth}{!}{

\begin{tabular}{|l|c|c|c|c|l|}
\hline
\multirow{2}{*}{\bf Model} & \multirow{2}{*}{\bf Method} & \multicolumn{2}{c|}{\sc { \bf Accuracy}} & \multirow{2}{*}{\bf Mem} & \multirow{2}{*}{\bf FLOPs}\\
\cline{3-4}
 &  & Top-1 & Top-5 &   &  \\
\hline
\multirow{5}{*}{AlexNet} & FPrec & 57.1\% & 80.2\% & 1x & 1135 (9.4x)\\
 & WBin (BWN) & 56.8\% & 79.4\% & 10.4x & 780 (6.4x)\\
 & FBin (XNOR) & 43.3\% & 68.4\% & 10.4x & {\bf 121 (1x)} \\
 & Hybrid-1 &  48.6\% & 72.1\% & 10.4x & 174 (1.4x)\\
 & Hybrid-2 & {\bf  48.2\%} & {\bf 71.9\%} & {\bf 31.6x} & 174 (1.4x)\\
\hline
Increase & Hybrid vs FBin & \textcolor{darkspringgreen}{+4.9\%} & \textcolor{darkspringgreen}{+3.5\%} & \textcolor{darkspringgreen}{+21.2x} & \textcolor{red}{+53 (+0.4x)} \\
\hline
\multirow{5}{*}{ResNet-18} & FPrec & 69.3\% & 89.2\% & 1x & 1814 (13.5x)\\
 & WBin (BWN) & 60.8\% & 83.0\% & 13.4x & 1030 (7.7x)\\
 & FBin (XNOR) & 51.2\% & 73.2\% & 13.4x & {\bf 134 (1x)}\\
 & Hybrid-1 & 54.9\% & 77.9\% & 13.4x & 359 (2.7x)\\
 & Hybrid-2 & {\bf 54.8\%} & {\bf 77.7\%} & {\bf 31.2x} & 359 (2.7x)\\
\hline
Increase & Hybrid vs FBin & \textcolor{darkspringgreen}{+3.6\%} & \textcolor{darkspringgreen}{+4.5\%} & \textcolor{darkspringgreen}{+17.8x} & \textcolor{red}{+225 (+1.7x)}\\
\hline
\end{tabular}}
\caption{Our hybrid models compared to FBin, WBin and NoBin models on Imagenet in terms of accuracy, memory and computations expense.}
\label{table:versions_imagenet}
\vspace{-0.4cm}
\end{table}
\begin{table}[t]
\resizebox{\columnwidth}{!}{
\begin{tabular}{|l|c|c|c|c|l|}
\hline
\multirow{2}{*}{\bf Model} & \multirow{2}{*}{\bf Method} & \multicolumn{2}{c|}{\sc { \bf Accuracy}} & \multirow{2}{*}{\bf Mem} & \multirow{2}{*}{\bf FLOPs}\\
\cline{3-4}
 &  & TU-Berlin & Sketchy &  &  \\
\hline
\multirow{4}{*}{Sketch-A-Net} & FPrec & 72.9\% & 85.9\% & 1x & 608 (7.8x)\\
 & WBin (BWN) & 73\% & 85.6\% & 29.2x & 406 (5.2x)\\
% & FBin (XNOR) & 48.2\% & 38.6\% & 29.2x & 78 (1x) \\
 & FBin (XNOR) & 59.6\% & 68.6\% & 19.7x & {\bf 78 (1x)} \\
 & Hybrid & {\bf 73.1\%} & {\bf 83.6\%} & {\bf 29.2x} & {\bf 85 (1.1x)}\\
\hline
Increase & Hybrid vs FBin & \textcolor{darkspringgreen}{+13.5\%} & \textcolor{darkspringgreen}{+15.0\%} & \textcolor{darkspringgreen}{+9.5x} & \textcolor{red}{+7 (+0.1x)}\\
\hline
\multirow{4}{*}{ResNet-18} & FPrec & 74.1\% & 88.7\% & 1x & 1814 (13.5x) \\
 & WBin (BWN) & 73.4\% & 89.3\% & 31.2x & 1030 (7.7x)\\
 & FBin (XNOR) & 68.8\% & 82.8\% & 31.2x & {\bf 134 (1x)} \\
 & Hybrid  & {\bf 73.8\%} & {\bf 87.9\%} & {\bf 31.2x} & 359 (2.7x)\\
\hline
Increase & Hybrid vs FBin & \textcolor{darkspringgreen}{+5.0\%} & \textcolor{darkspringgreen}{+5.1\%} & - & \textcolor{red}{+225 (+1.7x)}\\
\hline
\end{tabular}}
\caption{Our hybrid models compared to FBin, WBin and full prec models on TU-Berlin and Sketchy datasets in terms of accuracy, memory and speed tradeoff.} 

\label{table:tub_recacc}
\end{table}
On the TU-Berlin and Sketchy datasets in Table \ref{table:tub_recacc}, we find that Sketch-A-Net and ResNet-18 have significantly higher accuracies in the hybrid models compared to their FBin counterparts, a 13.5\% gain for Sketch-A-Net and 5.0\% for ResNet-18. 

These hybrid models also achieve over 29x compression over FPrec models and with a reasonable increase in the number of FLOPs - a mere 7M increase in Sketch-A-Net and a decent 225M increase in ResNet-18. We also compare them with state-of-the-art sketch classification models in Table \ref{table:sketchcomp}. Our hybrid Sketch-A-Net and ResNet-18 models achieve similar accuracies to state-of-the-art, while also highly compressing the models upto 233x compared to the AlexNet FPrec model. 
\begin{figure*}
\vspace*{-0.75cm}
\begin{center}
\resizebox{\textwidth}{!}{
\begin{tabular}{cc}
\includegraphics[scale=2]{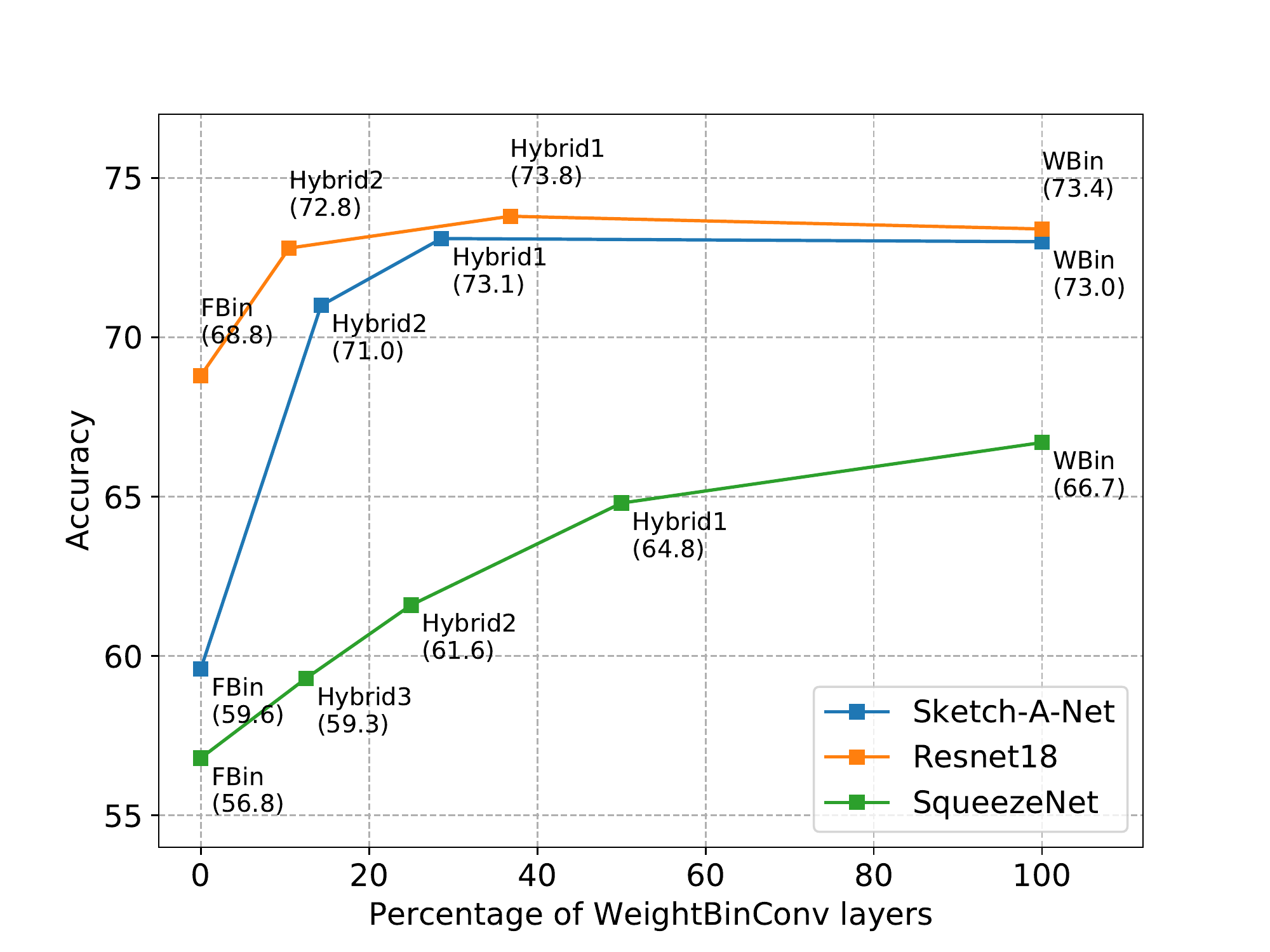} &
\includegraphics[scale=2]{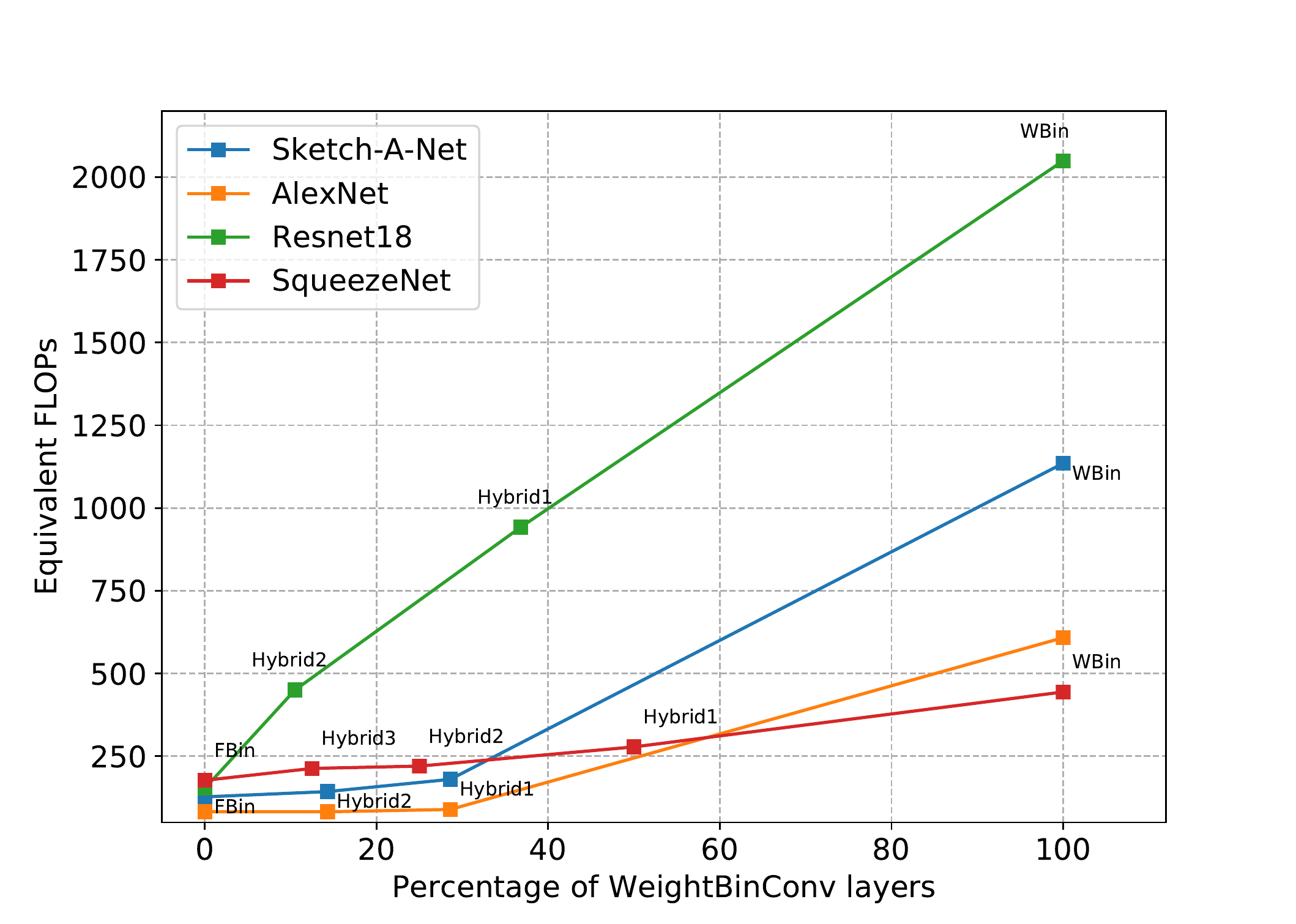}\\
\end{tabular}
}
\end{center}
\vspace*{-0.5cm}
\caption{Trade-off between WeightBinConv layers and accuracy on the TU-Berlin dataset is shown in the left figure, while the trade-off between weight binarized layers and speedup is shown in the right figure. Early on, we observe that a small increase in the percentage of WeightBinConv layers leads to a large increase in accuracy and a marginal decrease in speed. We achieve accuracies comparable to the WBin model with much fewer WeightBinConv layers.
}
\label{fig:tradeoff}
\vspace*{-0.3cm}
\end{figure*}
Thus, we find that our hybrid binarization technique finds a balance between sacrificing accuracy and gaining speedups and compression for various models on various datasets.
\begin{table}[t]
\begin{center}
\begin{tabular}{|l|c|c|c|l|}
\hline
{\bf Model} & {\bf Acc} & {\bf Mem} & {\bf FLOPs}\\
\hline
AlexNet-SVM & 67.1\% & 1x & 1135 (13.4x)\\
AlexNet-Sketch & 68.6\% & 1x & 1135 (13.4x)\\
Sketch-A-Net SC & 72.2\% & 8x & 608 (7.2x)\\
\hline
Sketch-A-Net-Hybrid & {\bf 73.1\%} & {\bf 233x} & {\bf 85 (1x)}\\
ResNet18-Hybrid & {\bf 73.8\%} & - & {\bf 359 }\\
\hline
Humans & {73.1\%} & - & - \\
Sketch-A-Net-2  \footnotemark \cite{yu2017sketch} & {\bf 77.0\%} & 8x & 608 (7.2x)\\
\hline
\end{tabular}
\footnotetext{It is the sketch-a-net SC model trained with additional imagenet data, additional data augmentation strategies and considering an ensemble, hence would not be a direct comparison}
\end{center}
\vspace{-0.5cm}
\caption{A comparison between state-of-the-art single model accuracies of recognition systems on the TU-Berlin dataset.}
\label{table:sketchcomp}
\vspace{-0.5cm}
\end{table}

\subsection{Algorithmic Insights}
We gained some insights into where to binarize from our investigation. We provide them as a set of practical guidelines to enable rapid prototyping of hybrid models, which gives meaningful insights into which layers were partitioned.

%without having to run the partitioning algorithm and give meaningful insights into the structure of binary networks.\\
{\bf Convert layers towards the end to WeightBinConv:} It is observed that later layers typically have high error rates, more filter repetitions, and lower computational cost. Hence, the algorithm tends to start converting models to Hybrid from the last layers. 

{\bf Convert the smaller of the layer placed parallely to WeightBinConv:} It is a good idea to convert the smaller of the parallely placed layers in the architecture like Residual layers in the ResNet architecture to WeightBinConv, since converting them to WeightBinConv would not damage the computational speedup obtained by the parallel FullBinConv layers.

{\bf Pick a low Hybridization Ratio:} Try to pick low values of the Hybridization Ratio $R$, ensuring a low proportion of number of layers the highest-error cluster.

{\bf Relax the Hybridization Ratio for compact models:} Having a higher Hybridization Ratio for compact models which inherently have fewer flops leaves more layer inputs un-binarized and retains accuracy. 

\subsection{Why are layer-wise errors independent?}
Can binarization noise introduced in a layer propagate further into the network and influence other layers? Hubara \etal \cite{hubara2016quantized} provide some insights for the same. Let $\mathbf{W}$ be the weight and $\mathbf{I}$ be the input to the convolutional layer. The output of the convolution between the binary weights and inputs can be represented by \begin{equation}\mathbf{O_B} = \alpha \cdot (sgn(\mathbf{W}^\intercal) \odot sgn(\mathbf{I}))\end{equation} The desired output $\mathbf{O}$ is modelled by $\mathbf{O_B}$ along with the binarization noise $\mathbf{N}$ introduced due to the function $sgn(.)$. \begin{equation}\mathbf{O} = \mathbf{W} * \mathbf{I} = \sum_{i}^{} \mathbf{O_B}_i + \mathbf{N_i}\end{equation} When the layer is wide, we expect the deterministic term $\mathbf{O_B}$ to dominate, because the noise term $\mathbf{N}$ is a summation over many independent binarizations from all the neurons in the layer. Thus, we argue that the binarization noise $\mathbf{N}$ should have minimal propagation and do little to influence the further inputs. Hence, it is a reasonable approximation to consider the error across each layer independently of the other layers.

\subsection{Variation with the Hybridization Ratio ($R$)}
To observe the trade-off between accuracy and speedup on different degrees of binarization, we chose different values of the Hybridization Ratio ($R$) to create multiple hybrid versions of the AlexNet, ResNet-18 and SqueezeNet models. Picking a larger $R$ would result in a higher number of WeightBinConv layers. We compare these hybrid networks to their corresponding FBin and WBin versions. 

In Figure \ref{fig:tradeoff}, we show model accuracies of AlexNet, ResNet-18 and SqueezeNet on the ImageNet dataset plotted against the number of WeightBinConv layers, starting from only FBin versions on the left, to only WBin versions on the right. We observe that in the case of AlexNet and ResNet-18, which are large models, we recover WBin accuracies quickly, at around the 35\% mark (Roughly a third of the network containing WeightBinConv layers), with low computational trade-off. We also observe that on sketch data, hybrid models tend to perform significantly better and perform on par with their WBin counterparts. 

We also notice that the smaller a model, the more trade-off must be made to achieve WBin accuracy, i.e a larger Hybridization Ratio must be used. AlexNet, the largest model crosses WBin accuracy at around 32\%, while ResNet-18, being smaller, saturates at around 40\%. SqueezeNet, a much more compact model, reaches its WBin accuracy at 60\%.

\subsection{Optimizing Memory}
We measured accuracies for FBin and Hybrid variants of Sketch-A-Net and ResNet-18 models on TU-Berlin and Sketchy Datasets with weights of the last layer binarized as well as non-binarized and the results are presented in Table \ref{table:otherresults}. For AlexNet-styled architectures (Sketch-A-Net), we observe a drastic drop in accuracies (From 59.1\% to 48.3\%) on binarizing the last layer, similar to observations made in previous binarization works \cite{zhou2016dorefa,tang2017train}. 

\begin{table}[t]
\resizebox{\columnwidth}{!}{
  \centering
\begin{tabular}{|l|c|c|c|c|c|}

\hline
{\bf Model} & {\bf BinType} & {\bf Last Bin?} & {\bf Acc } & {\bf Mem}\\
\hline
\multirow{2}{*}{Sketch-A-Net} & \multirow{2}{*}{FBin (XNOR)} & No & 59.6\% & 19.7x \\
 &  & Yes & 48.3\% & {\bf 29.2x} \\

\hline
\multirow{2}{*}{Sketch-A-Net} & \multirow{2}{*}{Hybrid} & No & 73.1\% & 19.7x \\
 &  & Yes & 72.0\% & {\bf 29.2x} \\
\hline
\multirow{2}{*}{Resnet-18} & \multirow{2}{*}{FBin (XNOR)} & No & 69.9\% & 13.4x \\
 &  & Yes & 68.8\% & {\bf 31.2x} \\

\hline
\multirow{2}{*}{Resnet-18} & \multirow{2}{*}{Hybrid} & No & 73.9\% & 13.4x \\
 &  & Yes & 73.8\% & {\bf 31.2x} \\
\hline
\end{tabular}}
\caption{Effects of last layer weight-binarization on TU-Berlin dataset, for Sketch-A-Net and ResNet-1. Observe that our hybrid models do not face drastic accuracy drop when the last layer is weight-binarized.}
\label{table:otherresults}
\vspace{-0.3cm}
\end{table}

Many efforts were made to quantize the last layer and avoid this drop. DoReFaNet and XNOR-Net did not binarize the last layer choosing to incur a degradation in model compression instead while \cite{tang2017train} proposed an additional scale layer to mitigate this effect. However, our hybrid versions are able to achieve similar accuracies (a 1\% drop for hybrid Sketch-A-Net and no drop for ResNet-18 or AlexNet) since the last layer is weight binarized instead. Hence, our method preserves the overall speedup even though we only weight-binarize the last layer, owing to the comparatively smaller number of computations that occur in this layer.

Note that the first layer is always a full-precision Conv layer. The reasons behind this are the insights obtained from \cite{anderson2017high}. They state that the first layer of the network functions are fundamentally different than the computations being
done in the rest of the network because the high variance principal components are not
randomly oriented relative to the binarization. Also, since it contains fewer parameters and low computational cost, it does not affect our experiments.

\subsection{Compressing Compact Models}
Whether compact models can be compressed further, or {\it need} all of the representational power afforded through dense floating-point values is an open question asked originally by \cite{iandola2016squeezenet}. 

We show that our hybrid-binarization technique can work in tandem with other compression techniques, which do not involve quantization of weights/activations and that hybrid binarization is possible even on compact models. We apply hybrid binarization to SqueezeNet\cite{iandola2016squeezenet} %and NIN-Net\cite{lin2013network}, 
a recent model that employed various architectural design strategies to achieve compactness. SqueezeNet achieves an 8x compression on the compact architecture of Sketch-A-Net. On applying hybrid binarization we achieve a further 32x compression, an overall 256x compression with merely 6\% decrease in accuracy. This is due to the high rate of compression inherent and further compression is difficult due to the small number of parameters. After showing that efficacy of hybrid binarization in the previous section, we show that hybrid binarization can work in combination with other compression techniques here.

\begin{table}[t]
\centering
\resizebox{\columnwidth}{!}{
\begin{tabular}{|l|c|c|c|c|l|}
 \hline
\multirow{2}{*}{\bf Model} & \multirow{2}{*}{\bf Method} & \multicolumn{2}{c|}{\sc { \bf Accuracy}} & \multirow{2}{*}{\bf Mem} & \multirow{2}{*}{\bf FLOPs}\\
\cline{3-4}
 &   & TU-Berlin & Sketchy &   &  \\
\hline
Sketch-A-Net & FPrec & 72.9\% & 85.9\% & 1x & 1135 (12.3x)\\
Squeezenet & FPrec & 71.2\% & 86.5\% & 8x & 610 (6.6x)\\
Squeezenet & WBin & 66.7\% & 81.1\% & 23.7x & 412 (4.5x)\\
\hline
Squeezenet & FBin & 56.8\% & 66.0\% & 23.7x & {\bf 92 (1x)} \\
Squeezenet & Hybrid & {\bf 64.8\%} & {\bf 79.6\%} & {\bf 23.7x} & 164 (1.8x)\\
\hline
Improvement & Hybrid vs FBin & \textcolor{darkspringgreen}{+8.0\%} & \textcolor{darkspringgreen}{+13.6\%} & - & \textcolor{red}{+72 (+0.8x)}\\
\hline
\end{tabular}}
\caption{Our performance on SqueezeNet, an explicitly compressed model architecture. Although SqueezeNet is an inherently compressed model, our method still achieves further compression on it.}
\label{table:squeezenet}
\vspace{-0.4cm}
\end{table}

Results for SqueezeNet are shown in Table \ref{table:squeezenet} for the TU-Berlin and Sketchy datasets, and we see that accuracy is only slightly lower compared to the hybridized versions of ResNet-18 and Sketch-A-Net on the same. Hybrid SqueezeNet achieves a total compression of 256x. Similarly, this technique can be combined with many techniques such as HWGQ-Net \cite{cai2017deep} which proposes an alternative layer to ReLU and repeated binarization as illustrated in \cite{tang2017train} among others. Since our primary goal is to investigate the viability of hybrid binarization, these investigations- albeit interesting, are out of the scope of our current work.

%------------------------------------------------------------------------
\section{Conclusion}
We proposed a novel algorithm for selective binarization of CNNs, which strikes a balance between performance, memory-savings and accuracy. The accuracies of our hybrid models were on par with their corresponding full-precision networks on TU-Berlin and Sketchy datasets, while providing the benefits of network binarization in terms of speedups, compression and energy efficiency. We successfully weight-binarized the last layers without significant accuracy drops, a problem faced by previous works in this area. We also showed that we can successfully combine the advantages of our approach with other architectural compression strategies, to obtain highly efficient models with negligible accuracy penalties. 

%We believe that the question of where to binarize a network is an interesting question and has potential for further investigation, both theoretically and practically (ANOOP!). 

\cleardoublepage
{\small
\bibliographystyle{ieee}
\bibliography{egbib}
}

\section{Appendix}
\appendix

\section{Introduction}
The supplementary material consists of the following:
\begin{enumerate}
\item Experimental details and hyperparameters used in our experiments can be found in the section- Experimental Details.
\item Tables containing parameters and FLOPs of individual layers of FBin, WBin and Hybrid models of AlexNet- along with the calculation formulae used and the percentage of unique filters.
\item Architecture diagrams of versions (FPrec, WBin, FBin, Hybrids) of Sketch-A-Net, ResNet-18, SqueezeNet, and AlexNet, where the reader can observe what Conv layers in the FPrec models have been replaced by FullBinConv and WeightBinConv layers on application of the proposed algorithm.
\end{enumerate}
\section{Experimental details}
\subsection{Data processing} 
For all the datasets, we resized the images to $256\times 256$. A $224\times 224$ ($225\times 225$  for Sketchanet) sized crop was randomly taken from an image with standard augmentations such as rotation and horizontal flipping for TU-Berlin and Sketchy. In the TU-Berlin dataset, we use three-fold cross-validation which gives us a 2:1 train-test split to make our results comparable to all previous methods. For Sketchy, we use the training images for retrieval as the training images for classification and validation images for retrieval as the validation images for classification. We train models using the standard training and validation data for ImageNet-12. We report ten-crop accuracies for TU-Berlin and Sketchy, and only single-crop accuracies for ImageNet.\\

\subsection{Hyper-parameters} 
We use the PyTorch framework to train our networks. We used TU-Berlin and Sketchy datasets to evaluate Sketch-A-Net, ResNet-18, and SqueezeNet (v1.0) architectures, and ImageNet data on AlexNet and ResNet-18 architectures. Each FullBinConv block was structured as in XNOR-Net (Batchnorm-Activ-Conv-ReLU). Each WeightBinConv and Conv block has the standard convolutional block structure (Conv-Batchnorm-ReLU). Weights of all layers excepting the first were binarized throughout our experiments, unless specified otherwise. We used a dropout of 0.2 before the last two convolutional layers in Sketch-A-Net and AlexNet, and a dropout of 0.2 before the last layer in SqueezeNet except after an FBin layer as followed in the XNOR-Net paper. All networks were trained from scratch. We used the Adam optimizer for all the models with a maximum learning rate of 0.002 and a minimum of 0.00005 with a decay factor of 2.  We do not use a bias term or weight decay for FullBinConv and WeightBinConv layers. We used a batch size of 256 for all Sketch-A-Net models and a batch size of 128 for ResNet-18 and SqueezeNet models, the maximum size which fits in a 1080Ti GPU.\\

\section{FLOPs, Exploiting Filter Repetition and Computational Cost Calculation}
Layers of AlexNet and ResNet-18 models are shown in following Tables, along with the number of parameters and the corresponding FLOPs. The number of parameters of each layer for FPrec and Binary versions of the model shown in multiples of 0.1 million for the sake of clarity. The number of parameters in the Binary version of a given layer (\#BinParams) is calculated as follows: \\
$$ \textnormal{\#BinParams} = \frac{\textnormal{\#Params}}{32}$$\\

FLOPs through each layer are given for FPrec, WBin, FBin and Hybrid versions, in multiples of 10 million. Except for the first layers, where weights are not binarized, the other layers have 58 times lesser FBin FLOP values due to enabling of XNOR/popcount operations. The number of repeated parameter values are also indicated and equivalent number of parameters in corresponding binarized layers are calculated as 
$$ \textnormal{\#WeightBinConvFLOPs} = \textnormal{\#FLOPs} \times (1-\textnormal{Repeated})$$
Then, we calculate the FLOPs in the FBin model by
$$\textnormal{\#FullBinConvFLOPs} = \frac{\textnormal{\#WbinFLOPs}}{58}$$
Further on, we select the parameters and FLOPs for each layer of the Hybrid models by selecting whether the layer is WeightBinConv or FullBinConv and pick the corresponding FLOPs.\\
The total number of parameters and FLOPs are calculated as the sum of individual layer FLOPs.

\section{Models used}
{\bf AlexNet:} A deep CNN that paved the way for most state-of-the-art CNN based models that exist today, and serves as a benchmark for comparison of network compression techniques.\\
{\bf ResNet-18:} A residual, implicitly compact model which is substantially deeper than previous standard models, that can be trained easily. \\
{\bf Sketch-A-Net:} A benchmark AlexNet-based model that was specifically engineered for sketches that beats human accuracy at sketch recognition, beating standard models fine-tuned on ImageNet which were unsuitable for sketch data. \\
{\bf SqueezeNet:} An explicitly compact CNN architecture achieving AlexNet accuracy with 50x fewer parameters.\\

\section{Model Architectures}
\begin{figure*}[t]
\resizebox{\textwidth}{!}{
\includegraphics[]{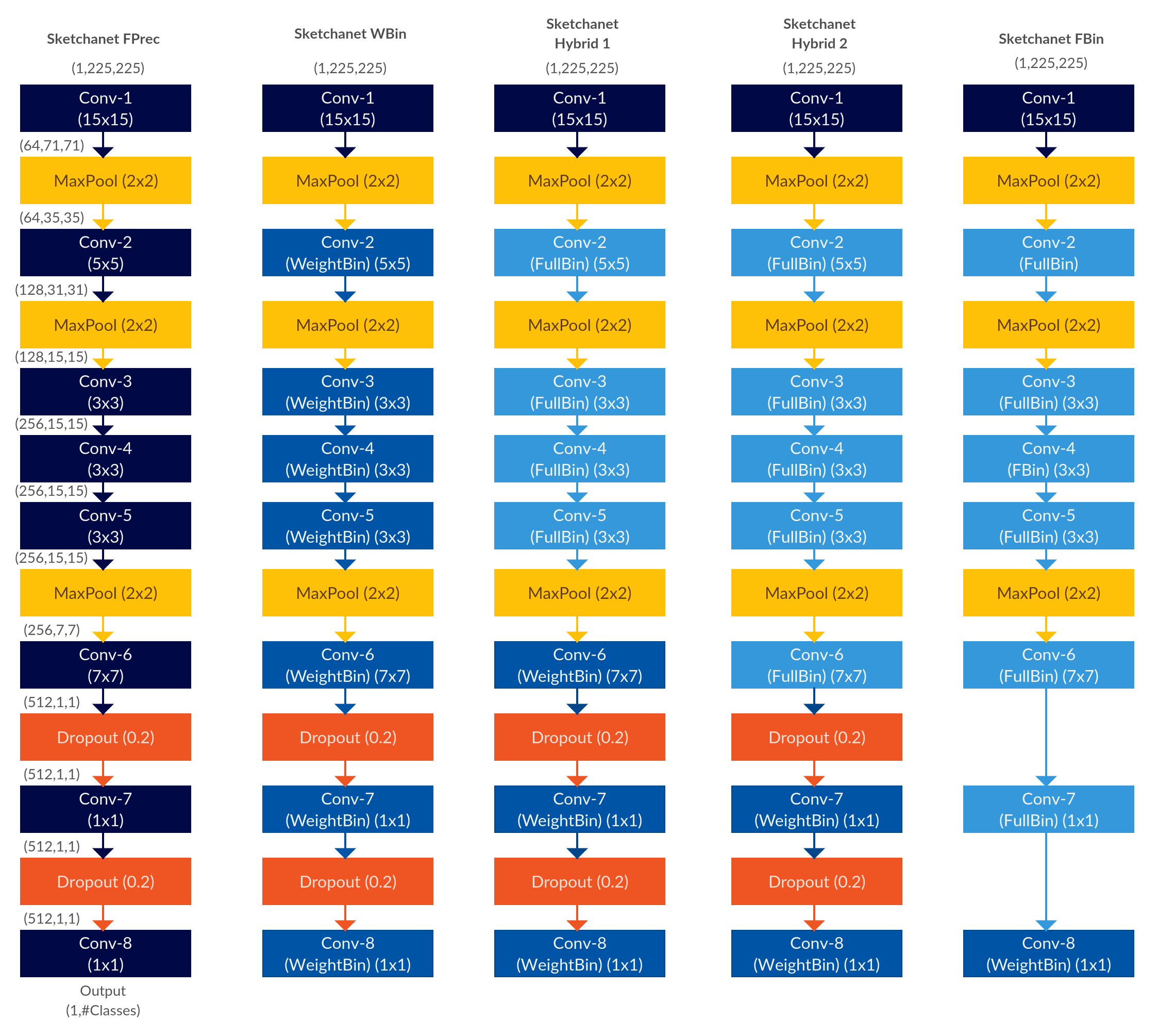}\\
}
\caption{Comparing architectures of FPrec, Wbin, Fbin, and two Hybrid versions of Sketch-A-Net. Our hybrid versions replace most conv layers with FullBinConv layers, but replace layers towards the end with WeightBinConv layers, following the algorithm.}
\vspace*{-0.5cm}
\label{fig:sketchanet}
\end{figure*}

\begin{figure*}[t]
\resizebox{\textwidth}{!}{
\includegraphics[]{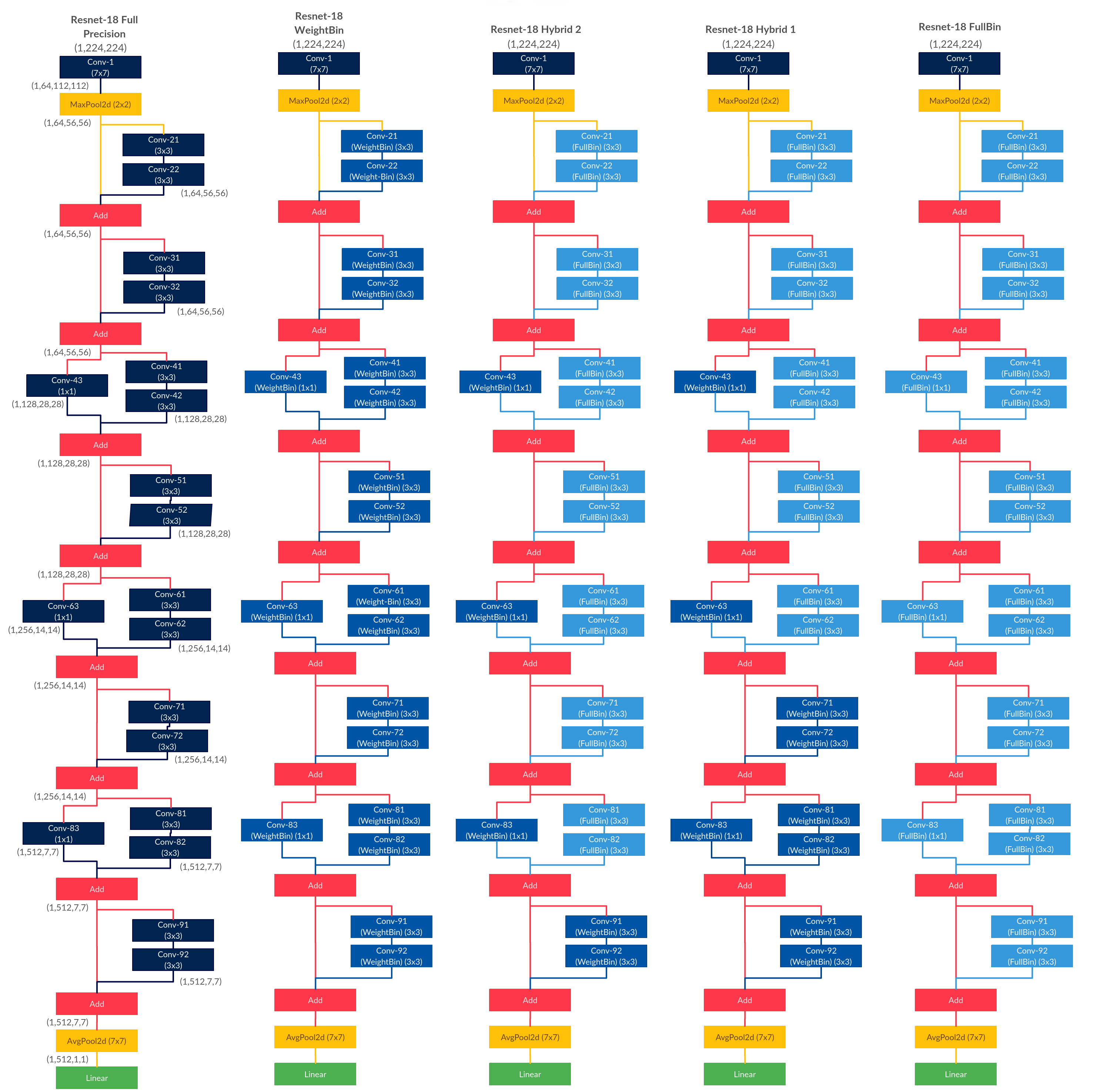}\\
}
\caption{Comparing architectures of FPrec, Wbin, Fbin, and two Hybrid versions of ResNet-18.}
\label{fig:resnet}
\end{figure*}

\begin{figure*}[t]
\resizebox{\textwidth}{!}{
\includegraphics[]{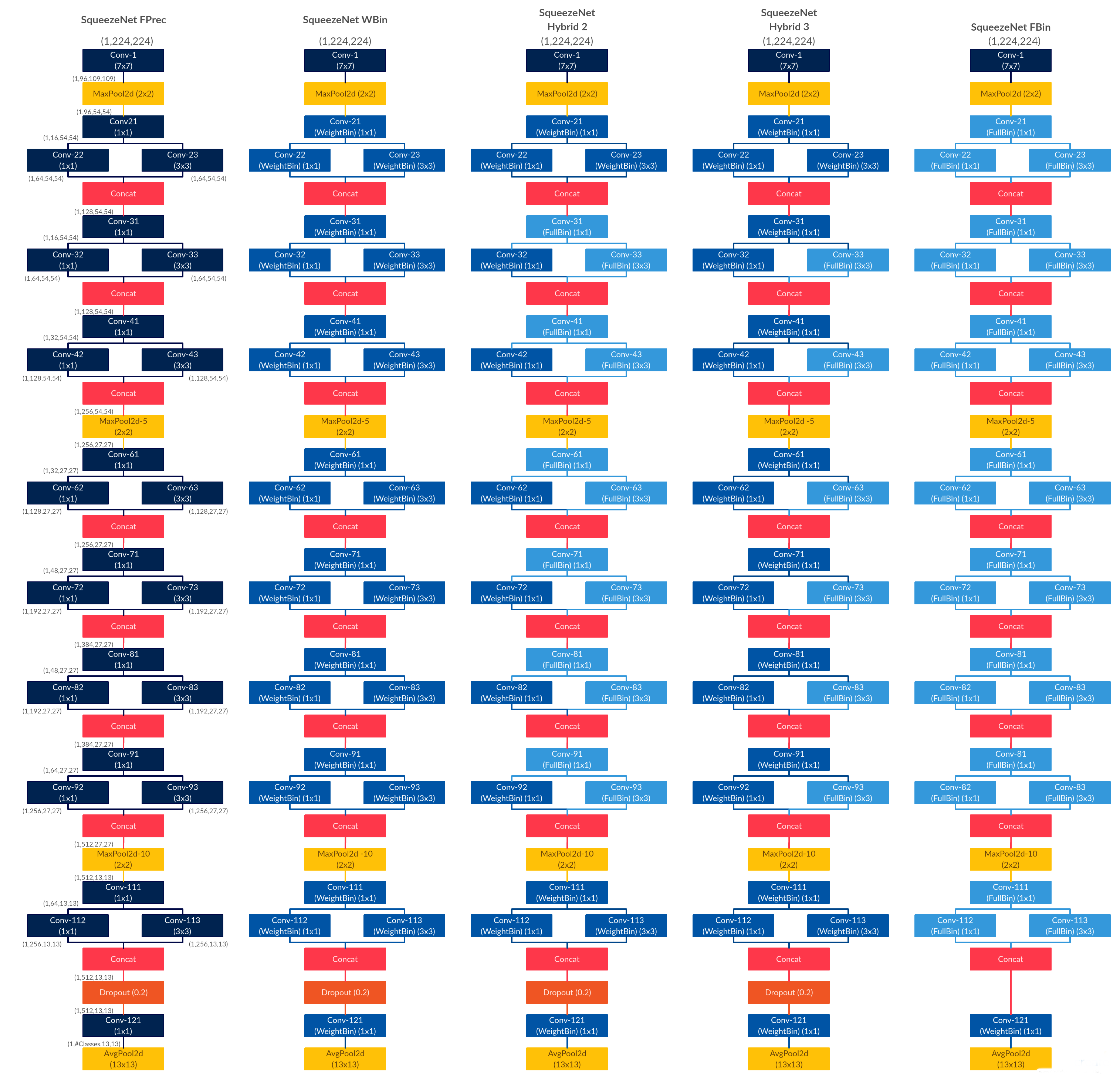}\\
}
\caption{Comparing architectures of FPrec, Wbin, Fbin, and two Hybrid versions of SqueezeNet.}
\label{fig:squeezenet}
\end{figure*}

Figure \ref{fig:sketchanet} is a comparison of architectures of FPrec, WBin, FBin, and two Hybrid versions of the Sketch-A-Net model. The hybrids replace almost all convolutional layers with FullBinConv layers, except the ones towards the end, which are replaced with WeightBinConv layers. Figures \ref{fig:resnet} and \ref{fig:squeezenet} show a similar architectural comparison for ResNet-18 and SqueezeNet models.  

\section{Binarization-errors across layers }
In Figure 2 in the main paper, we use the proposed metric to measure binarization-error across layers in Sketch-A-Net, ResNet-18, and SqueezeNet models. We use stars to indicate where our algorithm replaces a layer with a WeightBinConv layer, and squares when our algorithm replaces a layer with a FullBinConv layer. Our metric is a function of approximation-error and the inverse of the number of FLOPs through the layer. Observe that our algorithm replaces layers with high error scores with FullBinConv layers, which occur towards the end, and weight-binarizes the rest. \\
The partitioning algorithm clustered these layers to mark them for full-binarization and weight-binarization. For example, in ResNet-18, the algorithm gave two major partitions - layers 2 to 11 in the first, and 12 to 16 in the next, with a significant difference in the cluster means. Hence, layers 12 to 16 were binarized for the hybrid model.

\begin{table}[t]
\resizebox{\columnwidth}{!}{
\begin{tabular}{|l|c|c|c|c|c|c|c|c|}
\hline
{\bf Layers} &  \multicolumn{2}{c|}{\sc { \bf Parameters in 0.1M}} & \multicolumn{5}{c|}{\sc { \bf FLOPs }}\\
\hline
 & {\bf FPrec} & {\bf Bin} & {\bf FPrec} & {\bf Repeats} & {\bf WBin} & {\bf FBin} & {\bf Hybrid}\\
\hline
\multicolumn{8}{|c|}{\sc { \bf AlexNet}} \\
\hline
Conv1 & 0.23 & 0.232 & 10.54 & 0.00 & 10.54 & 10.54 & 10.54\\
Conv2 & 3.07 & 0.096 & 44.79 & 0 & 44.79 & 0.77 & 0.77\\
Conv3 & 6.64 & 0.207 & 14.95 & 0.71 & 4.34 & 0.07 & 0.07\\
Conv4 & 8.85 & 0.276 & 22.43 & 0.71 & 6.50 & 0.11 & 0.11\\
Conv5 & 5.90 & 0.184 & 14.95 & 0.60 & 5.98 & 0.10 & 0.10\\
Conv-FC1 & 377.49 & 11.796 & 3.77 & 0.00 & 3.77 & 0.07 & 3.77\\
Conv-FC2 & 167.77 & 5.243 & 1.68 & 0.00 & 1.68 & 0.03 & 1.68\\
Conv-FC3 & 40.96 & 1.280 & 0.41 & 0.00 & 0.41 & 0.41 & 0.41\\
Total & 610.90 & 19.316 & 113.53 &  & 78.01 & 12.11 & 17.47\\
\hline
\end{tabular}}
\caption{Layers of the AlexNet model, with the number of parameters and FLOPs for versions (WBin, Fbin, Hybrid, FPrec) of each. Also, the amount of unique parameters (a high number indicating high compressibility) is shown for each layer.}
\end{table}
\begin{table}[t]
\resizebox{\columnwidth}{!}{
\begin{tabular}{|l|c|c|c|c|c|c|c|c|}
\hline
{\bf Layers} &  \multicolumn{2}{c|}{\sc { \bf Parameters in 0.1M}} & \multicolumn{5}{c|}{\sc { \bf FLOPs }}\\
\hline
 & {\bf FPrec} & {\bf Bin} & {\bf FPrec} & {\bf Repeats} & {\bf WBin} & {\bf FBin} & {\bf Hybrid}\\
\hline
\multicolumn{8}{|c|}{\sc { \bf Sketch-A-Net}} \\
\hline
Conv1 & 0.23 & 0.232 & 7.26 & 0.00 & 7.26 & 7.26 & 7.26\\
Conv2 & 3.07 & 0.096 & 19.68 & 0.00 & 19.68 & 0.34 & 0.34\\
Conv3 & 6.64 & 0.207 & 6.64 & 0.58 & 2.80 & 0.05 & 0.05\\
Conv4 & 8.85 & 0.276 & 13.27 & 0.62 & 5.02 & 0.09 & 0.09\\
Conv5 & 5.90 & 0.184 & 13.27 & 0.61 & 5.19 & 0.09 & 0.09\\
Conv-FC1 & 47.19 & 1.475 & 0.64 & 0.00 & 0.64 & 0.01 & 0.64\\
Conv-FC2 & 2.62 & 0.082 & 0.03 & 0.00 & 0.03 & 0.00 & 0.03\\
Conv-FC3 & 1.28 & 0.040 & 0.05 & 0.00 & 0.05 & 0.05 & 0.05\\
Total & 75.77 & 2.593 & 60.84 &  & 40.68 & 7.89 & 8.54 \\
\hline
\end{tabular}}
\caption{Layer descriptions of the Sketch-A-Net model.}
\end{table}
\begin{table}[t]
\resizebox{\columnwidth}{!}{
\begin{tabular}{|l|c|c|c|c|c|c|c|c|}
\hline
{\bf Layers} &  \multicolumn{2}{c|}{\sc { \bf Parameters in 0.1M}} & \multicolumn{5}{c|}{\sc { \bf FLOPs }}\\
\hline
 & {\bf FPrec} & {\bf Bin} & {\bf FPrec} & {\bf Repeats} & {\bf WBin} & {\bf FBin} & {\bf Hybrid} \\
\hline
\multicolumn{8}{|c|}{\sc { \bf ResNet-18}} \\
\hline
Conv1 & 0.09 & 0.094 & 11.80 & 0.00 & 11.80 & 11.80 & 11.80\\
Conv2 & 0.37 & 0.012 & 11.56 & 0.23 & 8.86 & 0.15 & 0.15\\
Conv3 & 0.37 & 0.012 & 11.56 & 0.31 & 7.99 & 0.14 & 0.14\\
Conv4 & 0.37 & 0.012 & 11.56 & 0.23 & 8.90 & 0.15 & 0.15\\
Conv5 & 0.37 & 0.012 & 11.56 & 0.23 & 8.86 & 0.15 & 0.15\\
Conv6 & 0.74 & 0.023 & 5.78 & 0.49 & 2.92 & 0.05 & 0.05\\
Conv7 & 1.47 & 0.046 & 11.56 & 0.39 & 7.02 & 0.12 & 0.12\\
Conv8 & 0.08 & 0.003 & 0.64 & 0.00 & 0.64 & 0.01 & 0.01\\
Conv9 & 1.47 & 0.046 & 11.56 & 0.38 & 7.19 & 0.12 & 0.12\\
Conv2d & 1.47 & 0.046 & 11.56 & 0.39 & 7.08 & 0.12 & 0.12\\
Conv2d & 2.95 & 0.092 & 5.78 & 0.57 & 2.46 & 0.04 & 0.04\\
Conv2d & 5.90 & 0.184 & 11.56 & 0.50 & 5.74 & 0.10 & 0.10\\
Conv2d & 0.33 & 0.010 & 0.64 & 0.00 & 0.64 & 0.01 & 0.01\\
Conv2d & 5.90 & 0.184 & 11.56 & 0.52 & 5.51 & 0.10 & 5.51\\
Conv2d & 5.90 & 0.184 & 11.56 & 0.57 & 5.00 & 0.09 & 5.00\\
Conv2d & 11.80 & 0.369 & 5.78 & 0.70 & 1.76 & 0.03 & 1.76\\
Conv2d & 23.59 & 0.737 & 11.56 & 0.67 & 3.83 & 0.07 & 3.83\\
Conv2d & 1.31 & 0.041 & 0.64 & 0.00 & 0.64 & 0.01 & 0.64\\
Conv2d & 23.59 & 0.737 & 11.56 & 0.71 & 3.38 & 0.06 & 3.38\\
Conv2d & 23.59 & 0.737 & 11.56 & 0.76 & 2.73 & 0.05 & 2.73\\
Linear & 5.12 & 0.160 & 0.05 & 0.00 & 0.05 & 0.05 & 0.05\\
Total & 116.79 & 3.741 & 181.41 &  & 103.02 & 13.42 & 35.89\\
\hline
\end{tabular}}
\caption{Layers descriptions of the ResNet-18 model.}
\end{table}

\begin{table}[]
\resizebox{\columnwidth}{!}{
\begin{tabular}{|l|c|c|c|c|c|c|c|}
\hline
{\bf Layers} &  \multicolumn{2}{c|}{\sc { \bf Parameters in 0.1M}} & \multicolumn{5}{c|}{\sc { \bf FLOPs }}\\
\hline
 & {\bf FPrec} & {\bf Bin} & {\bf FPrec} & {\bf Repeats} & {\bf WBin} & {\bf FBin} & {\bf Hybrid} \\
\hline
\multicolumn{8}{|c|}{\sc { \bf Squeezenet}} \\
\hline
Conv2d & 0.14 & 0.141 & 16.77 & 0.00 & 16.77 & 16.77 & 16.77\\
Conv2d & 0.02 & 0.000 & 0.45 & 0.00 & 0.45 & 0.01 & 0.45\\
Conv2d & 0.01 & 0.000 & 0.30 & 0.00 & 0.30 & 0.01 & 0.30\\
Conv2d & 0.09 & 0.003 & 2.69 & 0.25 & 2.03 & 0.03 & 2.03\\
Conv2d & 0.02 & 0.001 & 0.60 & 0.00 & 0.60 & 0.01 & 0.60\\
Conv2d & 0.01 & 0.000 & 0.30 & 0.00 & 0.30 & 0.01 & 0.01\\
Conv2d & 0.09 & 0.003 & 2.69 & 0.20 & 2.15 & 0.04 & 0.04\\
Conv2d & 0.04 & 0.001 & 1.19 & 0.00 & 1.19 & 0.02 & 1.19\\
Conv2d & 0.04 & 0.001 & 1.19 & 0.00 & 1.19 & 0.02 & 0.02\\
Conv2d & 0.37 & 0.012 & 10.75 & 0.33 & 7.15 & 0.12 & 0.12\\
Conv2d & 0.08 & 0.003 & 0.60 & 0.00 & 0.60 & 0.01 & 0.60\\
Conv2d & 0.04 & 0.001 & 0.30 & 0.00 & 0.30 & 0.01 & 0.01\\
Conv2d & 0.37 & 0.012 & 2.69 & 0.36 & 1.73 & 0.03 & 0.03\\
Conv2d & 0.12 & 0.004 & 0.90 & 0.00 & 0.90 & 0.02 & 0.90\\
Conv2d & 0.09 & 0.003 & 0.67 & 0.00 & 0.67 & 0.01 & 0.01\\
Conv2d & 0.83 & 0.026 & 6.05 & 0.48 & 3.15 & 0.05 & 0.05\\
Conv2d & 0.18 & 0.006 & 1.34 & 0.00 & 1.34 & 0.02 & 1.34\\
Conv2d & 0.09 & 0.003 & 0.67 & 0.00 & 0.67 & 0.01 & 0.01\\
Conv2d & 0.83 & 0.026 & 6.05 & 0.54 & 2.77 & 0.05 & 0.05\\
Conv2d & 0.25 & 0.008 & 1.79 & 0.00 & 1.79 & 0.03 & 1.79\\
Conv2d & 0.16 & 0.005 & 1.19 & 0.00 & 1.19 & 0.02 & 0.02\\
Conv2d & 1.47 & 0.046 & 10.75 & 0.63 & 4.02 & 0.07 & 0.07\\
Conv2d & 0.33 & 0.010 & 0.55 & 0.00 & 0.55 & 0.01 & 0.55\\
Conv2d & 0.16 & 0.005 & 0.28 & 0.00 & 0.28 & 0.00 & 0.28\\
Conv2d & 1.47 & 0.046 & 2.49 & 0.74 & 0.66 & 0.01 & 0.66\\
Conv2d & 5.12 & 0.160 & 8.65 & 0.00 & 8.65 & 8.65 & 8.65\\
Total & 12.44 & 0.526 & 61.09 &  & 41.26 & 9.22 & 16.40\\
\hline
\end{tabular}}
\caption{Layers descriptions of the SqueezeNet model.}
\end{table}
\cleardoublepage

\begin{figure*}[t]
\resizebox{\textwidth}{!}{
\includegraphics[]{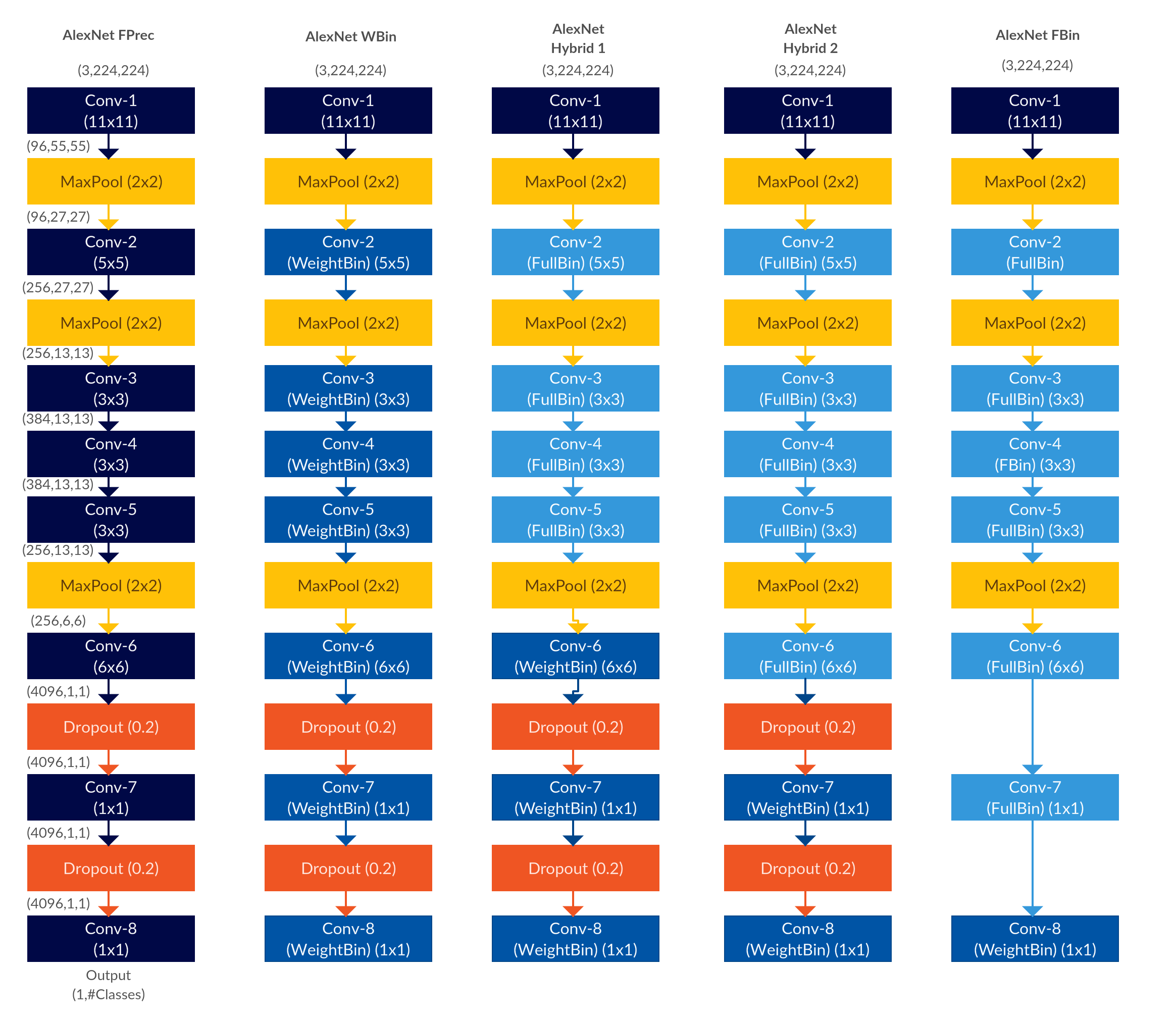}\\
}
\caption{Comparing architectures of FPrec, Wbin, Fbin, and two Hybrid versions of AlexNet.}
\label{fig:alexnet}
\end{figure*}

\end{document}